\documentclass[runningheads]{llncs}
\usepackage[T1]{fontenc}
\usepackage{graphicx,verbatim}
\usepackage{hyperref}
\usepackage{url}
\usepackage{booktabs}       
\usepackage{amsfonts}       
\usepackage{nicefrac}       
\usepackage{microtype}      
\usepackage{amsmath,bm}
\usepackage{amssymb}
\usepackage{array}
\usepackage{multirow}
\usepackage{makecell}
\usepackage{graphicx}
\usepackage{pifont}
\usepackage{marvosym}
\usepackage{adjustbox}
\usepackage{tabularx}
\usepackage{color}

\begin{document}
\title{Learning 3D Medical Image Models From Brain Functional Connectivity Network Supervision For Mental Disorder Diagnosis}
\titlerunning{Learning 3D Medical Image Models From FCNs for Diagnosis}

\author{Xingcan Hu\inst{1,2} \and Wei Wang\inst{1} \and Li Xiao\inst{1,2 (\text{\Letter})}}

\authorrunning{X. Hu et al.}

\institute{Department of Electronic Engineering and Information Science, \\
University of Science and Technology of China, Hefei~230052, China \and
Institute of Artificial Intelligence, Hefei Comprehensive National Science Center, Hefei~230088, China\\
\email{xiaoli11@ustc.edu.cn}\\
}

\maketitle              
\begin{abstract}
In MRI-based mental disorder diagnosis, most previous studies focus on functional connectivity network (FCN) derived from functional MRI (fMRI). However, the small size of annotated fMRI datasets restricts its wide application. Meanwhile, structural MRIs (sMRIs), such as 3D T1-weighted (T1w) MRI, which are commonly used and readily accessible in clinical settings, are often overlooked.
To integrate the complementary information from both function and structure for improved diagnostic accuracy, we propose CINP ({\bf C}ontrastive {\bf I}mage-{\bf N}etwork {\bf P}re-training), a framework that employs contrastive learning between sMRI and FCN. During pre-training, we incorporate masked image modeling and network-image matching to enhance visual representation learning and modality alignment.
Since the CINP facilitates knowledge transfer from FCN to sMRI, we introduce network prompting. It utilizes only sMRI from suspected patients and a small amount of FCNs from different patient classes for diagnosing mental disorders, which is practical in real-world clinical scenario.
The competitive performance on three mental disorder diagnosis tasks demonstrate the effectiveness of the CINP in integrating multimodal MRI information, as well as the potential of incorporating sMRI into clinical diagnosis using network prompting.


\end{abstract}
%
%
%
%
\section{Introduction}
By detecting the blood-oxygen-level-dependent (BOLD) responses to neural activity throughout the brain, functional MRI (fMRI) has become the leading neuroimaging technique for non-invasive study of human brain functions relevant to various behavioral and cognitive traits~\cite{logothetis2008we}. 
Recently, fMRI-derived functional connectivity network (FCN), as a graph architecture with nodes being brain regions-of-interest (ROIs) and each edge being functional connectivity (FC) between paired ROIs, has received considerable attention in diagnosis of mental disorders~\cite{yang2021alteration,bastos2016tutorial}, where FC is in general measured as statistical dependence between BOLD signals of paired ROIs.
So far a significant amount of work has been dedicated to learning deep and differentiable representations of FCN for improving diagnostic accuracy, such as graph neural networks (GNNs)~\cite{li2021braingnn,wang2024multiview}, convolutional neural networks (CNNs)~\cite{kawahara2017brainnetcnn}, and graph transformer~\cite{kan2022brain}.

Despite significant progress, such FCN-based deep learning methods for mental disorder diagnosis has yet to be widely adopted in real-world clinical practice. There are two pervasive challenges, i.e., the limited generalizability due to the insufficient annotated fMRI data volume, and the lack of integration with anatomical information from the easily obtainable structural MRI (sMRI), such as 3D T1w MRI. 
Since the anatomical structure of the brain inherently constrains its function~\cite{pang2023geometric}, 3D T1w MRI, which assesses brain anatomy, holds potential for diagnosing mental disorders. An efficient integration of structural and functional perspectives can provide a more comprehensive view of neurobiological abnormalities in mental disorders, leading to better diagnostic precision.

Motivated by the success of contrastive learning on large-scale image-text pairs~\cite{radford2021learning}, efforts in the biomedical domain have focused on pre-training vision-language models using medical images and their corresponding radiology reports~\cite{bannur2023learning}. Beyond images and texts, it is noteworthy that various MRI modalities, such as sMRI and fMRI, inherently provide contrasting perspectives by offering structural and functional information about the human brain, respectively.
For example, a bidrectional mapping scheme~\cite{ye2023bidirectional} performed contrastive learning between diffusion MRI-derived structural connectivity networks and BOLD signals.
F2TNet~\cite{he2024f2tnet} transferred knowledge from fMRI to sMRI using ROI-level contrastive learning, so as to enable accurate phenotypic predictions with sMRI alone.
For Alzheimer’s Disease prediction, Fedorov et al.~\cite{fedorov2024self} applied both inter- and intra-modal contrastive learning between sMRI and fALFF features. 
However, the scalability of these studies is somewhat limited by the specific model architecture, the small amount of data, and/or the coarse representation of features.

\begin{figure}[!t]
\includegraphics[width=0.85\textwidth]{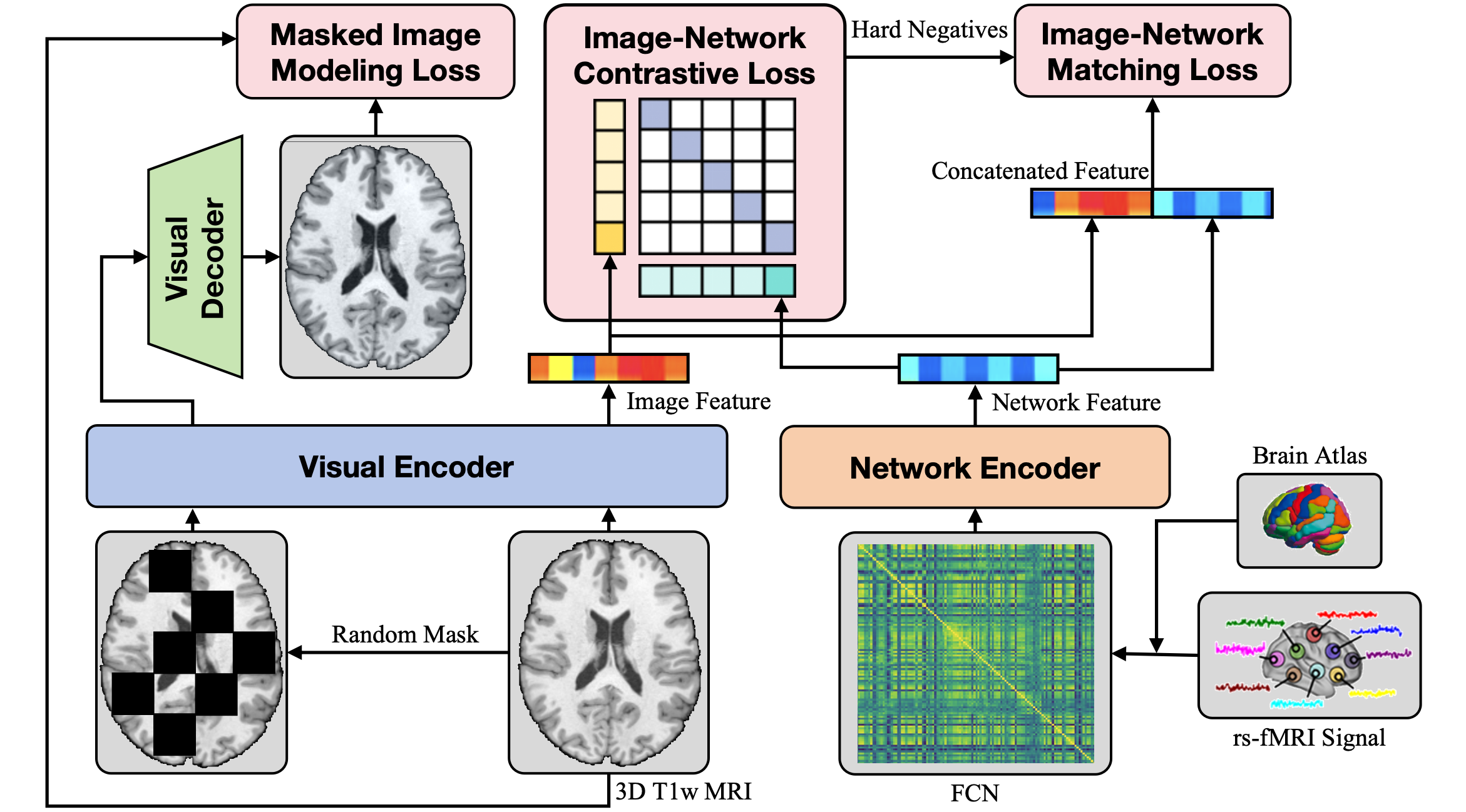}
\centering
\caption{
The framework of CINP. It primarily consists of a visual encoder, a visual decoder, and a network encoder.
}\label{fig1}
\end{figure}

In this regard, this paper focuses on scaling contrastive pre-training on suject-level sMRI and fMRI for mental disorder diagnosis. 
We collect a large cohort of $4619$ paired 3D T1w MRI images and fMRI-derived FCNs for pre-training. 
The Contrastive Image-Network Pre-training (CINP) framework (see Fig.~\ref{fig1}) is proposed to learn representations of 3D T1w MRI images through FCN supervision for mental disorder diagnosis tasks. Specifically, paired 3D T1w MRI images and FCNs are fed into a visual encoder and a network encoder to extract embeddings, respectively. The cosine similarity matrix between image embeddings and FCN embeddings is generated to compute image-network contrastive loss. Masked image modeling and image-network matching are used for better representation learning and modality alignment.
In particular, we develop a network prompting protocol, which leverages only 3D T1w MRI images from suspected patients and a small amount of FCNs from different patient classes for diagnosis of mental disorders. The similarities between the embedding of the 3D T1 MRI image of a suspected patient and the embeddings of FCNs are calculated. The patient is assigned to the class where the corresponding FCNs hold the highest similarity with the image.
The effectiveness of CINP is finally demonstrated on three mental disorder datasets by comparing CINP with FCN-based, sMRI-based, and multimodal models.

%
%
%
%
\section{Methods}

\subsection{Contrastive Image-Network Pre-training}
\label{object}
As illustrated in Fig.~\ref{fig1}, CINP is mainly composed of a visual encoder, a visual decoder, and a network encoder. 
For the visual encoder, an 8-layer 3D swin transformer is constructed and initialized with the weights from~\cite{tang2022self}. Given an input 3D T1w MRI image $I$, we randomly mask $30\%$ of the voxels, resulting in a masked MRI image $I^*$. Through the visual encoder, the normalized image embedding $\bm{v}_I$ and the normalized masked image embedding $\bm{v}_I^*$, both with a dimension of $768$, are obtained.
The visual decoder, based on the transpose convolution layer, follows the architecture in~\cite{tang2022self}, utilizing the masked image embedding $\bm{v}_I^*$ to reconstruct a volumetric MRI image $\hat{I}$.
We adopt brain network transformer (BNT)~\cite{kan2022brain} as the backbone for the network encoder, where an FCN $N$ is encoded into a $768$-dimensional normalized network embedding $\bm{w}_N$.

\subsubsection{Image-Network Contrastive Learning.}
Studies have shown that contrastive learning using image-text pairs can construct a joint semantic space of vision and language~\cite{radford2021learning}. Therefore, we aim to enhance the representations of 3D T1w MRI images with FCN supervision through contrastive learning between image-network pairs.
Specifically, given an image-network pair (i.e., a 3D T1w MRI image and an FCN), we aim to learn a similarity score $s(I,N) = \bm{v}_I^\mathsf{T} \bm{w}_N$, such that positive pairs (image and network from the same subject) have higher similarity scores, while negative pairs (image and network from different subjects) have lower similarity scores.
For each image and network in a batch, the softmax-normalized image-to-network and network-to-image similarities are calculated as
\begin{equation}
\label{contrastive_loss}
p_k^{\mathrm{in}}(I)=\frac{\mathrm{exp}(s(I,N_k)/\tau)}{\sum_{k=1}^K\mathrm{exp}(s(I,N_k)/\tau)} \ \ \text{and} \ \ \ p_k^{\mathrm{ni}}(N)=\frac{\mathrm{exp}(s(N,I_k)/\tau)}{\sum_{k=1}^K\mathrm{exp}(s(N,I_k)/\tau)},
\end{equation}
where the temperature factor $\tau$ is a learnable parameter and $K$ denotes the batch size. Let $\bm{y}^{\mathrm{in}}(I)$ and $\bm{y}^{\mathrm{ni}}(N)$ represent the ground-truth similarities of all images and networks, where positive pairs having a similarity of $1$ and negative pairs having a similarity of $0$.
Based on the cross entropy $\mathrm{H}(\cdot,\cdot)$, the image-network contrastive (INC) loss is defined as
\begin{equation}
\mathcal{L}_{\mathrm{INC}} = \frac{1}{2}\mathbb{E}_{(I,N)\sim D}\left[\mathrm{H}\left(\bm{y}^{\mathrm{ni}}(I),\bm{p}^{\mathrm{ni}}(I)\right)+\mathrm{H}\left(\bm{y}^{\mathrm{in}}(N),\bm{p}^{\mathrm{in}}(N)\right)\right].
\end{equation}

\subsubsection{Masked Image Modeling.}
Masked image modeling (MIM) aims to learn robust representations of MRI images. The MIM loss is defined by an $\mathrm{L_1}$ loss between the raw MRI image $I$ and the reconstructed MRI image $\hat{I}$, i.e.,
\begin{equation}
\mathcal{L}_{\mathrm{MIM}} = \mathbb{E}_{(I, \hat{I})\sim D}\lVert I-\hat{I}\rVert_1.
\end{equation}

\subsubsection{Image-Network Matching.}
Image-network matching (INM) is a binary classification task, which predicts whether a given image-network pair is from the same subject. Specifically, the concatenation of the image embedding $\bm{v}_I$ and the network embedding $\bm{w}_N$, denoted as $\left[\bm{v}_I,\bm{w}_N\right]$, is passed through a fully-connected layer to output a binary classification probability $q$. The INM loss is defined as
\begin{equation}
\mathcal{L}_{\mathrm{INM}} = \mathbb{E}_{(I,N)\sim D}\left[H\left(\bm{z}_{\mathrm{INM}},\bm{q}(I,N)\right)\right],
\end{equation}
where $\bm{z}_{\mathrm{INM}}$ is the ground-truth label represented by a 2-dimensional one-hot vector.
Moreover, we sample hard negatives as indicated in~\cite{li2021align}. To be specific, in each data batch, the image-network similarity in (\ref{contrastive_loss}) is used to sample image-network pairs which do not come from the same subject but own a high similarity for the INM loss. 
Finally, the complete pre-training loss of CINP is
\begin{equation}
\label{all_loss}
\mathcal{L} = \mathcal{L}_{\mathrm{INC}}+\alpha\mathcal{L}_{\mathrm{MIM}}+\beta\mathcal{L}_{\mathrm{INM}}.
\end{equation}

\subsection{Network prompting}

\begin{figure}[!t]
\includegraphics[width=0.9\textwidth]{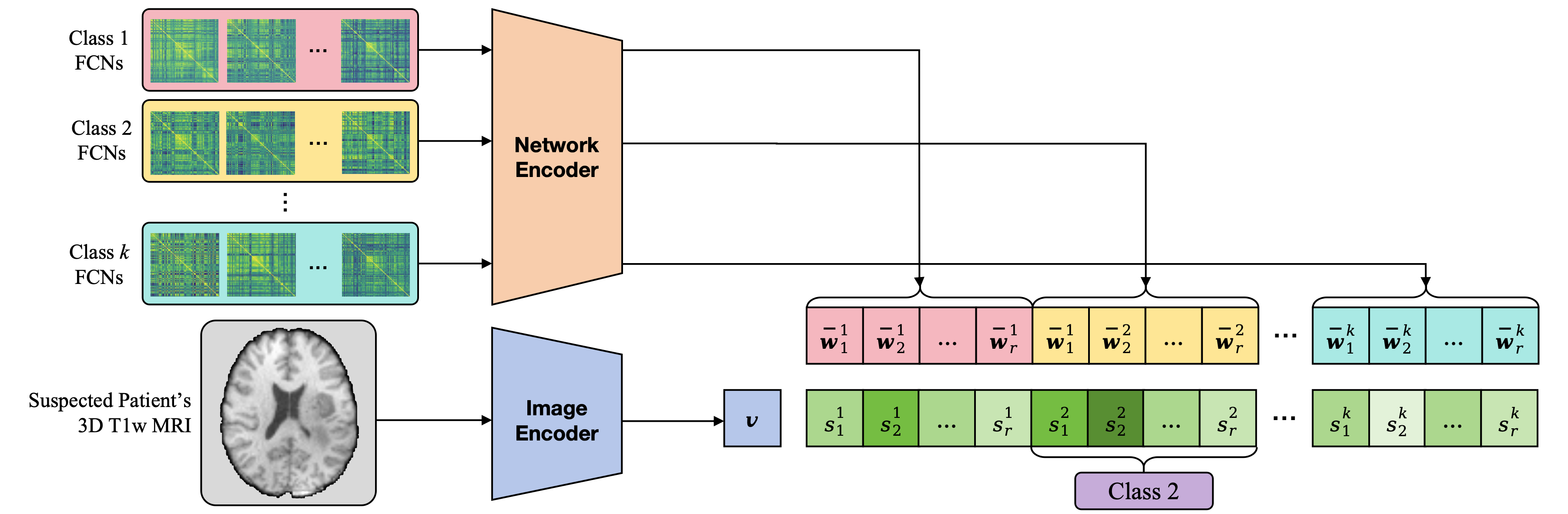}
\centering
\caption{
Workflow of network prompting. In this case, the 3D T1w MRI image is classified as class 2, since the mean of $s^2_1, s^2_2,\dots, s^2_r$ is the highest average similarity.
}\label{fig2}
\end{figure}

Previous methods employ linear probe or fine-tuning protocols to apply the pre-trained model to downstream tasks, which require a substantial number of annotated data and fall short when encountering the prevalent clinical scenario where fMRI are not routinely collected.
To address this challenge, as shown in Fig.~\ref{fig2}, we propose network prompting, inspired by the insight that FCNs from different subject classes (e.g., health controls and autism spectrum disorder patients) exhibit significant group-level differences~\cite{zhang2019strength,gratton2018functional}. We also hypothesize that the pre-trained CINP can measure the similarity between 3D T1w MRI image embeddings and FCN embeddings in the learned joint semantic space.

Specifically, we obtain a set of network embeddings $\mathcal{U}=\{\mathcal{C}_1,\mathcal{C}_2,\dots,\mathcal{C}_k\}$ of $k$ subject classes, each containing $n$ FCNs, i.e, $\mathcal{C}_l=\{\bm{w}^l_1,\bm{w}^l_2,\dots,\bm{w}^l_n\}$ for $l=1,2,\dots,k$. For each class, we partition the FCNs into $r$ disjoint subsets of equal size, i.e., $\mathcal{C}_l=\cup^r_{i=1}\mathcal{C}^i_l$, where $ \mathcal{C}^i_l\cap \mathcal{C}^j_l=\varnothing $ for $i\neq j$, and $|\mathcal{C}^i_l|=\dfrac{|\mathcal{C}_l|}{r}$ for $i=1,2,\dots,r$. 
The nework embeddings within the same subset are averaged to form $r$ group-level reference network embeddings:
$\overline{\mathcal{U}}=\{\{\bm{\overline{w}}^l_1,\bm{\overline{w}}^l_2,\dots,\bm{\overline{w}}^l_r,\}|l=1,2,\dots,k\}$, where $\bm{\overline{w}}^l_i=\dfrac{1}{|\mathcal{C}^i_l|}\sum_{\bm{w}\in \mathcal{C}^i_l}\bm{w}$ for $i=1,2,\dots,r$, which help eliminate subject-level biases.
Then, we calculate the similarities between the image embedding $\bm{v}$ of the suspected patient's 3D T1w MRI and the reference network embeddings, denoted as $\mathcal{S}=\{\{s^l_1,s^l_2,\dots,s^l_r\}|l=1,2,\cdots,k\}$, where $s^l_i=\bm{v}^\mathsf{T} \bm{\overline{w}}^l_i$ for $i=1,2,\dots,r$. 
Based on the average similarities of the image embedding with each class's FCNs $\overline{\mathcal{S}}=\{\overline{s}^1,\overline{s}^2,...,\overline{s}^k\}$, where $\overline{s}^l=\dfrac{1}{r}\sum_{i=1}^rs^l_i$ for $l=1,2,\dots,k$, we assign the patient to the class $k'$ with the highest average similarity, i.e., $\overline{s}^{k'}=\mathrm{max}(\overline{s}^1,\overline{s}^2,\dots,\overline{s}^k)$.

%
%
%
%
\section{Experiments}

\subsection{Experimental Settings}

\subsubsection{Datasets and Preprocessing.}
We collected several publicly available datasets where subjects have both 3D T1w MRI and resting-state fMRI (rs-fMRI) images. As a result, the large cohort for pre-training included four datasets (i.e., HBN~\cite{alexander2017open}, HCP~\cite{van2013wu}, QTIM~\cite{ds004169:1.0.7}, and CNP~\cite{ds000030:1.0.0}), while three mental disorder datasets (i.e., ABIDE~\cite{di2014autism}, ADHD~\cite{adhd2012adhd}, and SRPBS~\cite{tanaka2021multi}) were used for evaluation. We listed these datasets with basic demographic information in Table~\ref{datasets}. 
Using the fMRIPrep~\cite{esteban2019fmriprep} preprocessing pipeline, which is robust to variations in scan acquisition protocols, all MRI images were preprocessed and had a voxel size of $2\times2\times2 \ mm^{3}$.
To derive FCNs from rs-fMRI on the automated anatomical labelling (AAL) atlas~\cite{tzourio2002automated}, which contains $116$ ROIs, we calculated FC as Pearson's correlation between BOLD signals of paired ROIs. The corresponding row for each node in the FCN matrix was treated as the node features. 

\begin{table}[t]
  \caption{Demographic information of 7 datasets used in this study. (HC: Health Control, ASD: Autism Spectrum Disorder, ADHD: Attention Deficit Hyperactivity Disorder, MDD: Major Depression Disorder, SCZ: Schizophrenia)}
  \label{datasets}
  \centering
  \fontsize{8pt}{8pt}\selectfont 
  \setlength{\tabcolsep}{0.45mm}{
  \begin{tabular}{lccccc}
  
\toprule
Name                         & Usage                                  & Size & Gender (M/F) & Age (mean$\pm$sd) &  Samples          \\
\midrule
HBN~\cite{alexander2017open} & \multirow{4}*{\makecell{Pre-training}} & 2254 & 1455/799     & 10.73$\pm$3.39    & - \\
HCP~\cite{van2013wu}         & ~                                      & 1080 & 495/585      & (20-40)           & - \\
QTIM~\cite{ds004169:1.0.7}   & ~                                      & 1024 & 388/636      & 20.71$\pm$4.00    & - \\
CNP~\cite{ds000030:1.0.0}    & ~                                      & 261  & 152/109      & 33.29$\pm$9.29    & - \\
\midrule
ABIDE~\cite{di2014autism}    & \multirow{4}*{\makecell{Evaluation}}   & 855  & 719/136      & 16.92$\pm$7.91    & 395 ASD, 460 HC   \\
ADHD~\cite{adhd2012adhd}     & ~                                      & 872  & 538/334      & 11.98$\pm$3.34    & 325 ADHD, 547 HC  \\
SRPBS~\cite{tanaka2021multi} & ~                                      &1397  & 799/598      & 38.37$\pm$13.65   &
\makecell{125 ASD, 255 MDD, 147 SCZ, \\783 HC, 87 Others} \\
\bottomrule
  \end{tabular}}
\end{table}

\subsubsection{Implementation Details.}
\label{imple}
The CINP pre-training model was implemented with PyTorch 1.13.1 and MONAI 1.2.0.
For pre-training, we utilized the Adam optimizer with an initial learning rate of $10^{-5}$ and a weight decay of $10^{-5}$. The cosine annealing schedule was applied for the learning rate decreasing to $10^{-6}$. The batch size was set to $256$. The pre-training was performed on 8 NVIDIA A800 GPUs for $400$ epochs, taking approximately $100$ hours. Since the scales of three losses were similar, we set $\alpha=\beta=1$ in (\ref{all_loss}).
During the pre-training, the input 3D T1w MRI images were randomly augmented (i.e., Gaussian noise addition, flipping, intensity scaling and shifting) to learn robust representations. After augmentation, the volumes of all 3D T1 MRI images were resized to $96\times96\times96$.
Note that when evaluating CINP using the linear probe protocol, the input to the SVM classifier consisted solely of embeddings from 3D T1 MRI images.

\subsubsection{Performance Evaluation.}
The diagnosis of ASD and ADHD were treated as binary classification tasks on the ABIDE and ADHD datasets, respectively, and were evaluated using accuracy (ACC), area under the ROC curve (AUC), and Matthews correlation coefficient (MCC). For the SRPBS dataset, HC, ASD, MDD, SCZ, and other mental disorders were classified, where ACC and MCC were used for evaluation.
For linear probe and fine-tuning protocols, we randomly divided the evaluation dataset into training ($70\%$), validation ($10\%$), and testing ($20\%$) sets. Note that the network prompting protocol used the same data partitioning, but sampled only $10\%$ of FCNs from the training set to simulate a low-data scenario, while 3D T1w MRI images in the testing set were evaluated.

\subsection{Quantitative Results}

\subsubsection{Comparision with Baseline Models.}
We compared our CINP using the linear probe protocol with FCN-based, sMRI-based, and multimodal models.
Based on the implementation in the corresponding papers, four FCN-based and three multimodal models were deployed and trained from scratch on the three mental disorder datasets separately.
For three sMRI-based models using the linear probe protocol, we used the pre-trained weights provided in their respective papers. The feature maps from the last layer of these models were flattened, reshaped, and fed into an SVM classifier for classification. We also fine-tuned sMRI-based models for 10 epochs and presented their performances.

\paragraph{The FCN Supervision Enhanced the Embeddings of 3D T1 MRI Images.}
As shown in Table~\ref{metrics}, our CINP achieved the highest ACC on the ADHD and SRPBS datasets. 
Compared to the best results from sMRI-based and multimodal models, our CINP improved the ACC by $1.46\%$, $1.26\%$, and $1.21\%$ on the ABIDE, ADHD, and SRPBS datasets, respectively. It indicates that by conducting contrastive learning between 3D T1 MRI images and FCNs, their mutually complementary information can be fully captured, benefitting mental disorder diagnosis.

\paragraph{The Diagnostic Efficacy of MRI Modalities Differed by Mental Disorder.}
Although our CINP outperformed all competing models on the ADHD and SRPBS datasets, it did not achieve state-of-the-art performance on the ABIDE dataset. Meanwhile, FCN-based models performed worse than all other types of models on the ADHD dataset.
This suggests that sMRI is more effective in diagnosing ADHD, while ASD identification may require more information about brain function, even though the knowledge has been transferred from FCN to sMRI during contrastive pre-training.
Notably, on the SRPBS dataset, which included multiple mental disorders, multimodal models and CINP performed better, highlighting the importance of integrating sMRI and FCN for mental disorder subtyping.

\begin{table}[t]
  \caption{Classification results of different models on three mental disorder datasets.}
  \label{metrics}
  \centering
  \fontsize{8pt}{8pt}\selectfont 
  \begin{tabular}{cl|c|ccc|ccc|cc}
\toprule
\multirow{2}*{\makecell{Type}} & \multirow{2}*{\makecell{Method}} &  \multirow{2}*{\makecell{Training}} & \multicolumn{3}{c}{ABIDE}        & \multicolumn{3}{|c}{ADHD}    & \multicolumn{2}{|c}{SRPBS}           \\
\cmidrule(r){4-11}
~           & ~  & ~ & ACC & AUC & MCC & ACC & AUC & MCC & ACC & MCC\\

\midrule
\multirow{5}*{\makecell{FCN\\based}} & GCN~\cite{kipf2016semi} &  \multirow{5}*{\makecell{From \\scratch}} 
                     & 61.64 & 64.30 & 21.87 & 60.78 & 59.60 & 13.16 & 53.21 & 7.91  \\
~ & BrainGNN~\cite{li2021braingnn}             & ~ & 55.79 & 58.67 & 9.57  & 57.23 & 55.31 & 12.26 & 53.08 & 18.75 \\
~ & BrainNetCNN~\cite{kawahara2017brainnetcnn} & ~ & 62.92 & \underline{68.70} & 26.12 & 63.31 & 63.35 & 19.57 & 51.43 & 19.99 \\
~ & MHAHGEL~\cite{wang2024multiview}           & ~ & \underline{63.51} & 68.09 & \underline{29.19} & 64.11 & 62.00 & 18.41 & 56.58 & 20.22 \\
~ & BNT~\cite{kan2022brain}                    & ~ & \bf{65.96} & \bf{72.00} & \bf{31.80} & 63.42 & 64.47 & 19.86 & 57.08 & \bf{29.07} \\

\midrule
\multirow{6}*{\makecell{sMRI\\based}} & MedicalNet~\cite{chen2019med3d} &  \multirow{3}*{\makecell{Linear \\probe}} 
                   & 53.92 & 52.37 & 3.76  & 63.54  & 64.74 & 10.25 & 55.69 & 13.32  \\
~ &PRCLv2~\cite{zhou2023unified}      & ~ & 55.20 & 51.30 & 8.93  & 66.18  & 67.47 & 22.91 & 56.26 & 13.76  \\
~ &Swin-UNETR~\cite{tang2022self}  & ~ & 55.79 & 54.17 & 9.50  & 66.63  & 67.58 & 23.72 & 56.33 & 14.39  \\
\cmidrule(r){2-11}
~ & MedicalNet~\cite{chen2019med3d} & \multirow{3}*{\makecell{Fine \\tuning}}
                   & 54.39 & 51.25 & 4.56  & 65.71 & 68.38 & 24.09 & 58.57 & 14.10  \\
~ & PRCLv2~\cite{zhou2023unified}     & ~ & 54.60 & 59.65 & 14.19 & \underline{67.82} & 68.23 & 25.00 & 57.50 & 11.74 \\
~ & Swin-UNETR~\cite{tang2022self} & ~ & 57.31 & 59.83 & 14.53 & 67.39 & 68.39 & \bf{26.33} & 57.14 & 14.91  \\

\midrule
\multirow{3}*{\makecell{Multi\\modal}} & Cross-GNN~\cite{yang2023mapping} &  \multirow{3}*{\makecell{From \\scratch}} 
                                          & 61.40 & 62.43 & 23.04 & 65.52  & 66.14 & 24.74 & 60.22 & 21.09  \\
~ & MultiViT~\cite{bi2023multivit}    & ~ & 59.40 & 60.60 & 18.20 & 64.38  & 65.75 & 23.03 & \underline{63.08} & 22.18  \\
~ & CAMF~\cite{zhou2024interpretable} & ~ & 59.05 & 59.55 & 15.44 & 66.67  & \bf{71.56} & 24.21 & 61.29 & 22.10  \\

\midrule
\makecell{-} & CINP (Ours) & \makecell{Linear \\probe}
                    & 62.86 & 62.75 & 19.22 & \bf{69.08} & \underline{71.00} & \underline{25.33} & \bf{64.29} & \underline{22.26}\\
\bottomrule
  \end{tabular}
\end{table}

\subsubsection{Evaluation of Network Prompting.}
We evaluated the proposed network prompting protocol with different numbers of group-level reference networks ($r=1,5,10$) on the ABIDE and ADHD datasets. As shown in Table~\ref{np_results}, using only $10\%$ of the FCNs in the training set of evaluation datasets, the CINP with the network prompting protocol achieved the best MCC ($29.16\%$) on the ADHD dataset and outperformed all the sMRI-based and multimodal models on the ABIDE dataset. 
This demonstrates the feasibility of pre-training CINP with large-scale image-network pairs through contrastive learning and subsequently leveraging it with the network prompting protocol for mental disorder diagnosis, even when only a small number of FCNs from diagnosed patients are available.


\begin{table}[t]
  \caption{Classification results of the network prompting protocol with different numbers of group-level reference networks on two mental disorder datasets.}
  \label{np_results}
  \centering
  \fontsize{8pt}{8pt}\selectfont 
  \setlength{\tabcolsep}{2.5mm}{
  \begin{tabular}{c|ccc|ccc}
\toprule
\multirow{2}*{\makecell{Reference Network\\Number ($r$)}} & 
\multicolumn{3}{c}{ABIDE}        & 
\multicolumn{3}{|c}{ADHD}        \\
\cmidrule(r){2-7}
 ~  & ACC        & AUC        & MCC        & ACC        & AUC        & MCC        \\  \midrule
 1  & 56.45      & 58.74      & 15.09      & 62.27      & 61.78      & 21.70      \\
 5  & \bf{62.56} & \bf{61.34} & \bf{25.84} & 64.15      & \bf{65.33} & 27.11      \\
 10 & 57.24      & 56.10      & 13.97      & \bf{64.66} & 63.48      & \bf{29.16} \\
\bottomrule
  \end{tabular}}
\end{table}

\begin{table}[t]
  \caption{Classification results of CINP with different combinations of pre-training objective functions on three mental disorder datasets.}
  \label{ablation_table}
  \centering
  \fontsize{8pt}{7pt}\selectfont 
  \setlength{\tabcolsep}{1.2mm}{
  \begin{tabular}{ccc|ccc|ccc|cc}
    \toprule
\multicolumn{3}{c|}{Loss} & \multicolumn{3}{c|}{ABIDE}        & \multicolumn{3}{c|}{ADHD}    & \multicolumn{2}{c}{SRPBS}           \\
\cmidrule(r){4-11}
$\mathcal{L}_{\mathrm{INC}}$ & $\mathcal{L}_{\mathrm{MIM}}$ & $\mathcal{L}_{\mathrm{INM}}$
                                    &  ACC & AUC & MCC & ACC & AUC & MCC & ACC & MCC\\
\midrule
\ding{51} & \ding{55} & \ding{55}                  & 58.57 & 59.94 & 17.94 & 62.86 & 61.20 & 11.68 & 58.93 & 13.64 \\
\ding{51} & \ding{51} & \ding{55}          & 60.00 & 57.00 & 16.95 & \underline{66.23} & \underline{68.57} & \underline{20.67} & 60.71 & \underline{17.04} \\
\ding{51} & \ding{55} & \ding{51}          & \underline{61.42} & \underline{60.13} & \bf{21.52} & 63.37 & 64.69 & 17.50 & \underline{61.43} & 16.90                  \\
\ding{51} & \ding{51} & \ding{51} & \bf{62.86} & \bf{62.75} & \underline{19.22} & \bf{69.08} & \bf{71.00} & \bf{25.33} & \bf{64.29} & \bf{22.26}\\
\bottomrule
  \end{tabular}}
\end{table}

\subsection{Ablation Study}
We conducted ablation study on the CINP variants, which were pre-trained with different combinations of the three objective functions in Section~\ref{object}. As shown in Table~\ref{ablation_table}, both the MIM and INM losses improved the performance of CINP, with the best performance achieved when all three objective functions were used.
Specifically, on the ABIDE dataset, the MIM and INM losses improved the ACC by $1.44\%$ and $2.86\%$, respectively; on the ADHD dataset, the improvements were $5.71\%$ and $2.85\%$. 
Since the MIM loss primarily enhanced the representations of sMRI and the INM loss mainly transferred knowledge from FCNs to sMRI, this further implies that the diagnostic efficacy of MRI modalities varies across mental disorders from the objective function perspective.

\section{Conclusion}
In this paper, we proposed CINP, a framework that leverages contrastive learning between 3D T1 MRI and FCN. During pre-training, image-network contrastive loss, masked image modeling loss, and network-image matching loss were used to enhance the representations of 3D T1 MRI images with FCN supervision for downstream mental disorder diagnosis tasks.
With the pre-trained CINP, we introduced network prompting to utilize only sMRI from suspected patients and a small amount of FCNs from different patient classes for diagnosing mental disorders. Extensive experiments on three mental disorder classification tasks demonstrated the effectiveness of CINP, which sheds light on the potential of incorporating sMRI into clinical diagnosis.
%
%
%
%
\bibliographystyle{splncs04}
\bibliography{ref}

\end{document}